\documentclass[runningheads]{llncs}

\usepackage[T1]{fontenc}
\usepackage{subcaption} 
\usepackage{tikz}
\usepackage{siunitx}
\usepackage{booktabs}
\usepackage{amsmath}
\usepackage{xcolor}     
\usepackage{hyperref}   
\usepackage{graphicx}
\definecolor{lime}{HTML}{A6CE39}
\DeclareRobustCommand{\orcidicon}{%
	\begin{tikzpicture}
	\draw[lime, fill=lime] (0,0) 
	circle [radius=0.16] 
	node[white] {{\fontfamily{qag}\selectfont \tiny ID}};
	\draw[white, fill=white] (-0.0625,0.095) 
	circle [radius=0.007];
	\end{tikzpicture}
	\hspace{-2mm}
}

\foreach \x in {A, ..., Z}{%
	\expandafter\xdef\csname orcid\x\endcsname{\noexpand\href{https://orcid.org/\csname orcidauthor\x\endcsname}{\noexpand\orcidicon}}
}

\begin{document}
\title{Explainable Uncertainty Quantification for Wastewater Treatment Energy Prediction via Interval Type-2 Neuro-Fuzzy System}
\titlerunning{IT2-ANFIS for WWTP Energy Prediction}
%

\author{
Qusai Khaled\inst{1}{\orcidA{}} \and
Bahjat Mallak\inst{2} \and
Uzay Kaymak\inst{1}{\orcidB{}} \and
Laura Genga\inst{3}{\orcidC{}}
}

\authorrunning{Q. Khaled et al.}

\institute{
Jheronimus Academy of Data Science, Eindhoven University of Technology, Eindhoven, The Netherlands\\
\email{qusai.khaled@ieee.org, U.Kaymak@ieee.org}
\\[6pt]
\and
Haskoning, Amersfoort, The Netherlands\\
\email{bahjat.mallak@haskoning.com}  
\\[6pt]
\and
School of Industrial Engineering, Eindhoven University of Technology\\
\email{l.genga@tue.nl}
}
\maketitle
\begin{abstract}

Wastewater treatment plants consume 1–3\% of global electricity, making accurate energy forecasting critical for operational optimization and sustainability. While machine learning models provide point predictions, they lack explainable uncertainty quantification essential for risk-aware decision-making in safety-critical infrastructure. This study develops an Interval Type-2 Adaptive Neuro-Fuzzy Inference System (IT2-ANFIS) that generates interpretable prediction intervals through fuzzy rule structures. Unlike black-box probabilistic methods, the proposed framework decomposes uncertainty across three levels: feature-level, footprint of uncertainty identify which variables introduce ambiguity, rule-level analysis reveals confidence in local models, and instance-level intervals quantify overall prediction uncertainty. Validated on Melbourne Water's Eastern Treatment Plant dataset, IT2-ANFIS achieves comparable predictive performance to first order ANFIS with substantially reduced variance across training runs, while providing explainable uncertainty estimates that link prediction confidence directly to operational conditions and input variables.
\keywords{Wastewater Treatment \and Explainable AI \and Uncertainty Quantification \and Type-2 Fuzzy System }
\end{abstract}

\section{Introduction}

Wastewater treatment is a significant consumer of energy, accounting for roughly 1–3\% of total electricity consumption globally and even over 3\% in some regions like USA and UK \cite{gallo2024critical}. This substantial energy use makes wastewater treatment plants (WWTPs) critical targets for sustainability improvements, particularly as utilities face pressure to reduce operational costs and meet increasingly stringent environmental regulations \cite{capodaglio2025energy}. Predicting energy consumption can improve the sustainability of WWTP operations \cite{alali2023unlocking}. Energy forecasts allow plant operators to anticipate fluctuations in daily demand, optimize equipment scheduling, and align high-energy activities with renewable power availability \cite{amin2025characterization}. Such predictions support both immediate operational decisions and long-term planning for infrastructure expansion \cite{allen2025knowledge}. In the context of tightening climate targets, energy prediction has become an essential tool to align WWTP operations with carbon neutrality goals \cite{shao2024low}. However, these facilities exhibit substantial variability arising from influent characteristics (e.g., Biological Oxygen Demand (BOD) and Chemical Oxygen Demand (COD)), operational conditions (e.g., aeration and membrane performance), and environmental drivers (e.g., seasonal and temperature effects) \cite{yang2025application}. Fundamentally, the process characteristics are such that very accurate predictions are not possible due to large variation and lack of information. Hence, the models should quantify prediction uncertainty as well as make point predictions. As such, before transforming predictive modeling insights into engineering decision-making context in water treatment operation, there should be methods to evaluate the uncertainty of the model results \cite{belia2009wastewater}. Additional layer of complexity lies in the trustworthiness in these models, despite advances in artificial intelligence, its implementation in water treatment remains limited due to a lack of trust compared to knowledge-based models \cite{richards2023rewards}. Which is partially attributed to the lack of explainability of these models and the use of black box predictive models. Although extensive research is invested in the application of probability methods to perform uncertainty quantification, such as Bayesian neural networks and Gaussian processes \cite{williams2006gaussian}. The explainability of these models presents a significant challenge \cite{wang2023semantic}. Such methods produce statistical uncertainty estimates that cannot be directly linked to specific process variables or operational rules. This opacity creates barriers to trust and adoption in an industry where understanding causal relationships is essential for a trustworthy application. Limited work is done on the intersection of explainability and uncertainty quantification \cite{schiller2022explaining}

To address these challenges, this study presents a three-layered framework for predicting the energy consumption of a WWTP with explainable uncertainty quantification; instead of presenting uncertainty estimates by confidence interval, we train interval type-2 neuro-fuzzy inference system (IT2-ANFIS), then leverage the learned upper and lower bounds of membership functions to explain uncertainty in three levels, feature attribution level, local model level, and prediction instance level. The remainder of this paper is organized as follows. Section 2 reviews related work on energy consumption in WWTPs and IT2-ANFIS. Section 3 describes the proposed framework, Section 4 explains the experimental design, Section 5 presents the results and Section 6 concludes the paper.

\section{Background}

Recent WWTP advancements include real-time BOD-based aeration control, biogas generation achieving 50\% energy sufficiency, renewable energy integration, and IoT automation for monitoring \cite{guo2019integration}. Yet optimal energy efficiency remains challenging due to operational complexity and inherent uncertainty. Energy consumption varies with plant size, automation level, treatment technology, and influent characteristics \cite{longo2016monitoring}. Pumping and aeration account for the majority of WWTP electricity demand \cite{drewnowski2019aeration}. Aeration alone consumes 45-75\% in activated sludge systems, while pumping represents the second-largest demand. Energy patterns exhibit complexity due to fluctuating influent flows, variable pollutant loads, weather impacts, and non-linear relationships between biological process performance (affected by temperature and biomass activity) and energy requirements. Predictive models are critical for cost savings, optimization, regulatory compliance, and carbon reduction \cite{alali2023unlocking}. Machine learning approaches show promise, Bagehrzadeh et al. \cite{bagehrzadeh2021full} deployed ANN, LSTM, GBM, and Random Forest using climate, hydraulic, and wastewater parameters selected via mutual information, achieving 68 MWh Mean Absolute Error (MAE). Other studies explored K-nearest neighbor \cite{alali2023unlocking} and Random Forest \cite{zhang2021novel}. Li et al. \cite{li2019analysis} combined fuzzy clustering with radial basis function neural networks for energy-per-cubic-meter prediction. Deep learning addresses data scarcity: Oliveira et al. \cite{oliveira2021forecasting} applied transfer learning with LSTM, GRU, and CNN (with CNN optimal for advance forecasts), while Harrou et al. \cite{harrou2023energy} used cubic spline data augmentation with LSTM and Bi-GRU, achieving 1.36-2.18\% mean absolte percentage error (MAPE). Despite these advances, limited research provides WWTP energy models with uncertainty quantification capabilities.

Fuzzy systems offer strong potential for uncertainty quantification because they explicitly model vagueness and imprecision using fuzzy sets and graded membership functions rather than assuming crisp boundaries. Unlike black-box models, fuzzy systems operate through transparent rule-based structures that expose their reasoning mechanisms, establishing them as interpretable white-box or grey-box models within the explainable AI community \cite{ali2023explainable}. Various fuzzy architectures exist: Mamdani systems use fuzzy outputs, Takagi--Sugeno--Kang (TSK) systems employ crisp consequent functions enabling efficient optimization, while probabilistic fuzzy systems integrate statistical uncertainty. Type-2 fuzzy systems \cite{mendel2002type} extend classical type-1 fuzzy logic by introducing uncertainty in the membership functions themselves.  Type-2 fuzzy systems \cite{mendel2002type} extend classical (Type-1) fuzzy logic by introducing uncertainty in the membership functions. Interval Type-2 (IT2) systems represent this uncertainty using a Footprint of Uncertainty (FOU), defined by lower and upper membership functions that bound the interval of possible membership grades. This property explicitly captures epistemic uncertainty arising from imprecise data, measurement noise, or expert disagreement \cite{wu2007uncertainty}.

Applications of IT2 fuzzy systems for uncertainty quantification span diverse domains. In renewable energy forecasting, IT2 systems have successfully modeled solar radiation prediction uncertainty, with type-2 models achieving 1-3.7\% MAE improvement over type-1 systems while handling meteorological uncertainties \cite{almaraashi2024practical}. For wind power and load forecasting in microgrids, IT2-based prediction intervals captured generation uncertainties with specified coverage probabilities \cite{marin2016prediction}. In electric load forecasting, IT2 fuzzy systems with optimal type reduction algorithms outperformed traditional methods by explicitly handling demand uncertainties \cite{khosravi2013load}. Medical diagnosis applications leveraged IT2 uncertainty handling capabilities for heart disease \cite{nguyen2015medical}, and blood pressure load classification \cite{guzman2019design}, with studies showing improved diagnostic accuracy compared to type-1 systems under uncertain clinical data conditions \cite{ontiveros2020comparative}. Financial forecasting applications utilized IT2 fuzzy systems for stock market prediction \cite{jiang2018interval} \cite{bernardo2012interval}, with interval-valued outputs providing both point forecasts and uncertainty bounds that outperformed type-1 models in volatile market conditions \cite{takahashi2021new}. Despite extensive applications across these domains, limited work have explored application within the context of WWTPs, motivating the present work.

\section{Model Architecture and Training Methodology}

\begin{figure}[!htbp]
    \centering
    \includegraphics[width=0.9\columnwidth]{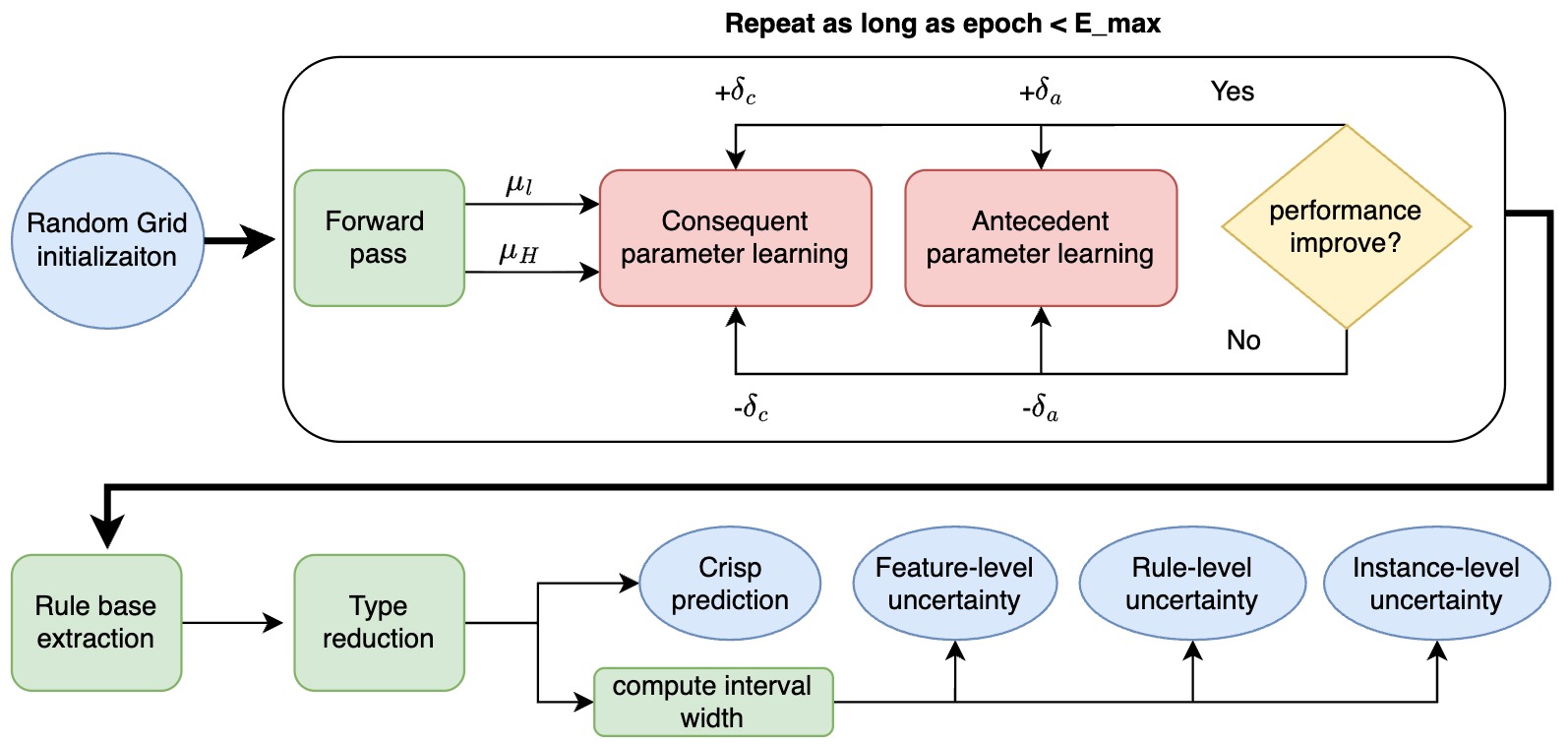}
    \caption{IT2-ANFIS Training Framework with Uncertainty Quantification}
    \label{fig:Frama}
\end{figure}

This section presents the IT2-ANFIS training framework. Figure~\ref{fig:Frama} shows the overall architecture, from rule initialization to uncertainty quantification. The model employs a first-order TSK system. Each rule is expressed as
\begin{equation}
\text{IF } x_1 \text{ is } \tilde{A}_1^j \text{ AND } \dots \text{ AND } x_F \text{ is } \tilde{A}_F^j, \text{ THEN } y_j = \mathbf{w}_j^T \mathbf{x} + b_j,
\end{equation}
where \( \tilde{A}_f^j \) is an interval type-2 Gaussian membership function with uncertain mean \([c1_f^j, c2_f^j]\) and standard deviation \(\sigma_f^j\). The consequent \(y_j\) is a linear combination of inputs, learned via mini-batch gradient descent (MBGD). The framework has three main steps. First, rule parameters \([c1_f^j, c2_f^j]\), \(\sigma_f^j\), \(\mathbf{w}_j\), and \(b_j\) are initialized. Second, membership estimation computes lower and upper membership functions, forming FOU. Third, type reduction produces a crisp prediction and quantifies three levels of uncertainty via a weighting factor \(q\).

\vspace{-5mm}
\vspace{-1cm}

\subsection{Random grid initialization}
For energy prediction in WWTP, considering hydraulic, operational, and climatological parameters would cause the input dimensionality to exceed 10 variables, making uniform grid partitioning \cite{khaled2025interpretable} prone to rule explosion (e.g., $2^{10} = 1024$ rules for 2 partitions per dimension). To address this, a random grid initialization strategy based on Latin Hypercube Sampling (LHS) \cite{mckay2000comparison} is employed to initialize cluster centers. This stratified sampling method ensures uniform coverage of the multidimensional input space: each feature range is divided into equiprobable intervals, from which points are sampled to produce a representative distribution of initial rule centers while avoiding rule explosion. The initialization process proceeds as follows:

\begin{enumerate}

    \item \textbf{LHS-based center selection}: 
    For each rule $j$ and feature $f$, the feature range $[x_f^{min}, x_f^{max}]$ is divided into $R$ equal segments, where $R$ is the number of rules. The rule center is then sampled as

    \vspace{-3mm}
    {\footnotesize
    \begin{equation}
    c_{jf} = x_f^{min} + j \cdot \frac{x_f^{max} - x_f^{min}}{R} + \mathcal{U}\left(0, \frac{x_f^{max} - x_f^{min}}{R}\right),
    \end{equation}
    }
    
    where $\mathcal{U}(a,b)$ denotes uniform random sampling from the interval $[a,b]$. This ensures that each rule occupies a distinct region of the input space while preserving randomness within its segment.  
    
    \item \textbf{Uncertain Mean Interval}: 
    Once the rule centers are initialized, the uncertain mean interval for each Gaussian membership function is defined by its lower and upper bounds:
    
    \vspace{-4mm}
    {\footnotesize
    \begin{align}
    c1_{jf} &= c_{jf} - \frac{\alpha \cdot w_f}{2}, &\quad
    c2_{jf} &= c_{jf} + \frac{\alpha \cdot w_f}{2}
    \end{align}
    }

    where $c_{jf}$ is the initial center point initialized via LHS, $c1_{jf}$ and $c2_{jf}$ denote the lower and upper bounds of the uncertain mean interval. The parameter $w_f$ is the estimated partition width of the feature $f$, defined as
    
    \vspace{-3mm}
    {\footnotesize
    \begin{equation}
    w_f = \frac{x_f^{max} - x_f^{min}}{\lceil R^{1/F} \rceil}.
    \end{equation}
    }
    
    Here $F$ is the number of input features, while $x_f^{min}$ and $x_f^{max}$ denote the minimum and maximum values of feature $f$. The ceiling function $\lceil \cdot \rceil$ ensures that partitioning remains consistent even when $R^{1/F}$ is not an integer. In this way, each rule is associated with an uncertain mean interval $[c1_{jf}, c2_{jf}]$ around its center $c_{jf}$. The parameter $\alpha$ is the uncertainty factor that controls how far these bounds initially extend around the center. These bounds define the admissible range within which the mean can vary and are optimized during training.

    \item \textbf{Standard deviation initialization}: The standard deviations $\sigma_{jf}$ that define the width of the Gaussian membership functions are initialized as:
    \vspace{-1mm}
    {\small
    \begin{equation}
    \sigma_{jf} = \max(0.5 \cdot w_f, \sigma_{min}),
    \end{equation}
    }
    where $\sigma_{min} = 0.05$ is the minimum allowable standard deviation to prevent overly narrow membership functions that could lead to numerical instability.

    \item \textbf{Consequent initialization}: To prevent gradient explosion in early training phases, $\mathbf{w}_j$ and $b_j$ are initialized with small random values drawn from a zero-mean normal distribution with standard deviation 0.01, i.e., $\mathcal{N}(0, 0.01^2)$.

\end{enumerate}

\subsection{Membership Functions and Firing Strengths}
The Footprint of Uncertainty (FOU) is the region bounded by lower and upper membership functions $\mu_L^{jf}(x_f)$ and $\mu_U^{jf}(x_f)$ and estimated as
{\small
\begin{equation}
\mu_L^{jf}(x_f) = \begin{cases}
\exp\left(-\frac{1}{2}\left(\frac{x_f - c2_{jf}}{\sigma_{jf}}\right)^2\right) & \text{if } x_f \leq c_{mid,jf} \\
\exp\left(-\frac{1}{2}\left(\frac{x_f - c1_{jf}}{\sigma_{jf}}\right)^2\right) & \text{if } x_f > c_{mid,jf}
\end{cases}
\end{equation}
}
{\small
\begin{equation}
\mu_U^{jf}(x_f) = \begin{cases}
\exp\left(-\frac{1}{2}\left(\frac{x_f - c1_{jf}}{\sigma_{jf}}\right)^2\right) & \text{if } x_f < c1_{jf} \\
1.0 & \text{if } c1_{jf} \leq x_f \leq c2_{jf} \\
\exp\left(-\frac{1}{2}\left(\frac{x_f - c2_{jf}}{\sigma_{jf}}\right)^2\right) & \text{if } x_f > c2_{jf}
\end{cases}
\end{equation}
}

where $c_{mid,jf} = \tfrac{c1_{jf} + c2_{jf}}{2}$ is the midpoint of the uncertain mean interval, computed dynamically from the current values of $c1_{jf}$ and $c2_{jf}$. Since $c1_{jf}$ and $c2_{jf}$ are optimized independently during training, $c_{mid,jf}$ shifts accordingly at each update step. From the membership functions, firing strengths are computed using the product t-norm

\vspace{-4mm}
{\small
\begin{align}
\mu_L^j(\mathbf{x}) &= \prod_{f=1}^{F} \mu_L^{jf}(x_f), &\quad
\mu_U^j(\mathbf{x}) &= \prod_{f=1}^{F} \mu_U^{jf}(x_f),
\end{align}
}

\noindent
Then, they are normalized as:
\vspace{-3mm}
{\small
\begin{align}
\bar{f}_L^j &= \frac{\mu_L^j(\mathbf{x})}{\sum_{k=1}^{R} \mu_L^k(\mathbf{x})}, &\quad
\bar{f}_U^j &= \frac{\mu_U^j(\mathbf{x})}{\sum_{k=1}^{R} \mu_U^k(\mathbf{x})}.
\end{align}
}

\subsection{Type Reduction and Output Computation}

Type reduction is the process of converting the interval type-2 fuzzy output into a type-1 fuzzy set or crisp value. In interval type-2 fuzzy systems, each input produces two outputs: a lower bound $y_{lower}$ and an upper bound $y_{upper}$, corresponding to the lower and upper membership functions, respectively. These bounds represent the uncertainty in the prediction. The lower and upper outputs are computed as
\vspace{-4mm}
{\small
\begin{align}
y_{lower} &= \sum_{j=1}^{R} \bar{f}_L^j \cdot y_j = \sum_{j=1}^{R} \bar{f}_L^j \cdot (\mathbf{w}_j^T \mathbf{x} + b_j), \\
y_{upper} &= \sum_{j=1}^{R} \bar{f}_U^j \cdot y_j = \sum_{j=1}^{R} \bar{f}_U^j \cdot (\mathbf{w}_j^T \mathbf{x} + b_j).
\end{align}
}

\noindent
Type reduction is performed using a design factor $q$, which produces the final crisp output as a weighted combination:
\vspace{-1mm}
{\small
\begin{equation}
y_{pred} = q \cdot y_{lower} + (1-q) \cdot y_{upper}
\end{equation},
}

\noindent

the q factor can be treated as either a fixed value or a learnable parameter. Allowing the model to optimize q during training enables it to adaptively balance between the lower and upper bounds of the interval type-2 output. A learned value of q > 0.5 indicates that the model places greater emphasis on the lower bound, reflecting a tendency toward conservative estimates and suggesting that the data or learned rules carry higher perceived uncertainty. Conversely, a learned value of q < 0.5 implies that the model relies more on the upper bound, yielding optimistic predictions and indicating that the training data exhibits lower variability.

\subsection{Optimization Strategy}
To balance sparsity and stability, consequent parameters ($\mathbf{w}_j, b_j$) are updated with combined L1 Lasso and L2 Ridge regularization:
\vspace{-1mm}
{\small
\begin{align}
\mathbf{w}_j^{(t+1)} &= \mathbf{w}_j^{(t)} - \eta_{cons} \left[\nabla_{\mathbf{w}_j}\mathcal{L}_{MSE} + \lambda_{L2} \mathbf{w}_j + \lambda_{L1} \cdot \text{sign}(\mathbf{w}_j)\right] \\
b_j^{(t+1)} &= b_j^{(t)} - \eta_{cons} \left[\nabla_{b_j}\mathcal{L}_{MSE} + \lambda_{L2} b_j + \lambda_{L1} \cdot \text{sign}(b_j)\right]
\end{align}
}

\noindent
Here, $\eta_{cons}$ denotes the consequent learning rate: an $L_1$ penalty with $\lambda_{L1} = 0.05$, which promotes sparsity in the consequent weights and thus supports implicit feature selection, and an $L_2$ penalty with $\lambda_{L2} = 0.001$, which provides weight decay to prevent overfitting. The gradient of the MSE loss function with respect to these parameters is computed using the normalized firing strengths:
\vspace{-2mm}
{\small
\begin{align}
\nabla_{\mathbf{w}_j}\mathcal{L} &= \frac{1}{N}\sum_{n=1}^{N} e_n \cdot \phi_n^j \cdot \mathbf{x}_n, &\quad
\nabla_{b_j}\mathcal{L} &= \frac{1}{N}\sum_{n=1}^{N} e_n \cdot \phi_n^j,
\end{align}
}

\vspace{-2mm}
\noindent
where $e_n = y_{pred}^n - y_{true}^n$ is the prediction error, and $\phi_n^j$ is the combined normalized firing strength factor
\vspace{-4mm}
{\small
\begin{equation}
\phi_n^j = \frac{q \cdot \bar{f}_L^j}{\sum_{k=1}^{R} \bar{f}_L^k} + \frac{(1-q) \cdot \bar{f}_U^j}{\sum_{k=1}^{R} \bar{f}_U^k}.
\end{equation}
}

\noindent
Antecedent parameters ($c1_{jf}, c2_{jf}$) are updated using PyTorch automatic differentiation to compute analytical gradients of the MSE loss function with respect to the FOU bounds. This approach replaces numerical finite-difference methods with exact gradient computation, which provides higher accuracy in gradient estimation, improved computational efficiency through vectorized operations, and greater numerical stability due to the use of automatic differentiation. Consequently, the update rule for the antecedent parameters is defined as:
{\small
\begin{align}
c1_{jf}^{(t+1)} &= c1_{jf}^{(t)} - \eta_{ant} \cdot \text{clip}\left(\nabla_{c1_{jf}}\mathcal{L}, -0.1, 0.1\right), \\
c2_{jf}^{(t+1)} &= c2_{jf}^{(t)} - \eta_{ant} \cdot \text{clip}\left(\nabla_{c2_{jf}}\mathcal{L}, -0.1, 0.1\right).
\end{align}
}
Here, the gradients $\nabla_{c1_{jf}}\mathcal{L}$ and $\nabla_{c2_{jf}}\mathcal{L}$ are computed on the full training set using PyTorch's backward pass through the entire IT2-TSK inference chain. The clipping operation applied in the update rule (i.e., restricting gradients within $[-0.1, 0.1]$) serves as gradient clipping, which is introduced to enhance stability during training and prevent excessively large parameter updates. After each update, constraints are enforced to maintain physical consistency of the antecedent parameters. Specifically, an ordering constraint ensures that $c1_{jf} \leq c2_{jf}$, with values swapped if violated, and a minimum separation constraint enforces $c2_{jf} - c1_{jf} \geq 0.05$ to prevent collapse of the footprint of uncertainty. In addition, multiple regularization techniques are incorporated into the implementation pipeline to prevent overfitting and ensure training stability as depicted in Table 1.
{\footnotesize
\renewcommand{\arraystretch}{1.3}
\vspace{-1cm}
\begin{table}[!htbp]
\centering
\scriptsize
\caption{Regularization techniques applied in IT2-ANFIS}
\label{tab:regularization}
\begin{tabular}{lll}
\hline
\textbf{Regularization Type} & \textbf{Target Parameters} & \textbf{Purpose} \\
\hline
L1 (Lasso) & Consequent weights \& biases & Feature selection, sparsity \\
L2 (Ridge) & Consequent weights \& biases & Weight decay, prevent overfitting \\
Gradient Clipping & Antecedent bounds ($c1, c2$) & Prevent large updates, training stability \\
Early Stopping & All parameters & Prevent overfitting \\
Model Checkpointing & All parameters & Restore best validation performance \\
\hline
\end{tabular}
\end{table}
}
\vspace{-0.5cm}

\section{Experimental settings}
\begin{figure}[!htbp]
    \centering
    \includegraphics[width=1.0\columnwidth]{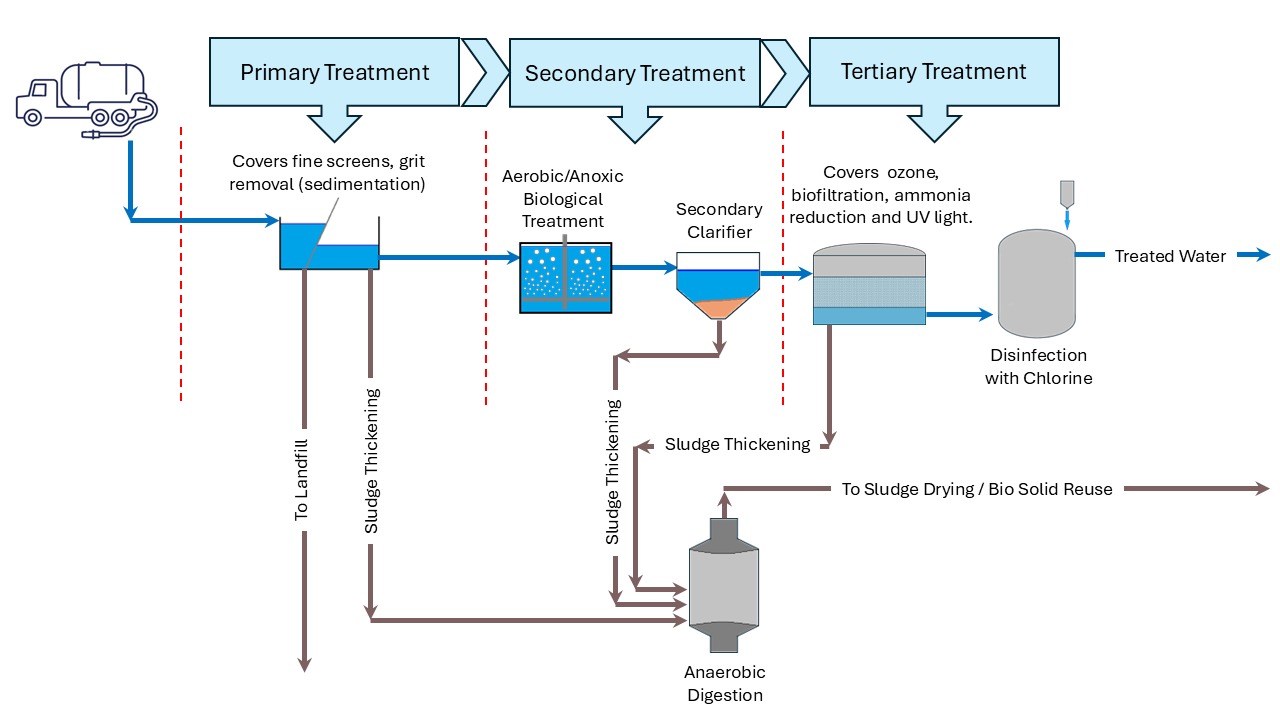}
    \caption{Simplified schematic of the wastewater treatment process at
    Melbourne Water's Eastern Treatment Plant.}
    \label{fig:scheme}
\end{figure}

This study utilizes an operational dataset combining operational data from Melbourne Water's Eastern Treatment Plant with meteorological observations from Melbourne Airport weather station, spanning approximately six years \cite{bagehrzadeh2021full}. The openly available dataset comprises 1,000 daily records, integrating energy consumption (MWh), hydraulic parameters, wastewater characteristics, and climate variables. A schematic view of the WWTP is shown in Figure \ref{fig:scheme}, which illustrates the main processing stages of the wastewater treatment plant as a flow-based system. Influent wastewater enters from the left and passes through three sequential modules: primary treatment (physical removal of solids), secondary treatment (biological removal of dissolved pollutants), and tertiary treatment (advanced polishing and disinfection). In parallel, the by-products of these stages (sludge) are routed to thickening and anaerobic digestion for stabilization and reuse. Hydraulic parameters include average inflow and outflow rates, which directly influence pumping energy requirements and treatment capacity utilization. Wastewater characteristics encompass ammonia, total nitrogen, BOD and COD. These biological parameters are critical drivers of energy consumption, as nitrogen removal through nitrification-denitrification processes is typically the most energy-intensive operation in wastewater treatment, requiring substantial aeration energy for biological oxidation. Similarly, high BOD and COD levels necessitate extended aeration periods and increased oxygen transfer, directly elevating electricity demand. Climate variables include average, maximum, and minimum temperatures, atmospheric pressure, average relative humidity, total precipitation, average visibility and wind speed. Temperature particularly affects biological process kinetics and equipment efficiency, while precipitation influences inflow volumes and dilution effects.

With respect to hyperparameters, the model employs an adaptive learning–rate mechanism that adjusts the update magnitudes according to the change in training error between successive iterations. Let $\Delta_{\mathrm{error}} = \mathrm{MSE}^{(t-1)} - \mathrm{MSE}^{(t)}$. Whenever the training error decreases ($\Delta_{\mathrm{error}} > 0$), the learning rates for both the antecedent and consequent parameters are incrementally increased by a factor of 1.05 to accelerate convergence. Conversely, when no improvement is observed ($\Delta_{\mathrm{error}} \leq 0$), the learning rate is conservatively reduced in order to stabilize training; the consequent learning rate is scaled by 0.9, while the antecedent learning rate is scaled by 0.95, reflecting their differing sensitivities. To ensure numerical stability, each learning rate is constrained within predefined bounds, specifically $\eta_{\mathrm{cons}} \in [10^{-5},\,0.05]$ and $\eta_{\mathrm{ant}} \in [10^{-6},\,0.02]$. The lower learning rate cap for antecedents reflects their greater sensitivity that have been observed empirically during the experiments. The uncertainty factor $\alpha$ was set to 0.2, corresponding to an initial uncertainty interval spanning 20\% of the estimated partition width for each input feature. This choice reflects a moderate level of antecedent uncertainty: large enough to capture ambiguity in the positioning of linguistic terms, yet sufficiently constrained to preserve separation between neighboring fuzzy rules and avoid excessive overlap. Empirically, values of $\alpha$ in the range [0.1, 0.3] were observed to provide stable training behavior, while larger values led to overly diffuse rule activation and reduced interpretability. Fixing $\alpha$ = 0.2 thus represents a balanced trade-off between uncertainty modeling, numerical stability, and semantic clarity. Type reduction factor $q = 0.5$ is fixed throughout training to maintain balanced contribution from both bounds of the type-2 fuzzy output. This design choice prevents bias toward either conservative (lower) or optimistic (upper) estimates and simplifies the type reduction process compared to iterative methods like the Karnik-Mendel algorithm \cite{wu2008enhanced}. By fixing $q$, we ensure consistent interpretation of prediction intervals while maintaining computational efficiency during training. Regarding evaluation, the dataset was split into training (64\%), validation (16\%), and test (20\%) sets, with model selection based on validation mean squared error (MSE). For comparative analysis, a systematic parameter sweep across 5--50 fuzzy rules in which IT2-ANFIS performance is compared against classical type-1 Adaptive Neuro-Fuzzy System (ANFIS). In addition, Random Forest (RF) and Support Vector Machines (SVM) are trained to enrich the baseline comparison landscape. Performance is reported as the mean across 10 independent training runs initiated at different random seeds. The full set of hyperparameters is provided in the accompanying code: \url{https://github.com/QusaiKhaled/XUQ}.
\section{Results and Discussion}
\subsection{Model selection and rule-complexity analysis}

\begin{figure}[!b]
    \centering
    \includegraphics[width=1.0\columnwidth]{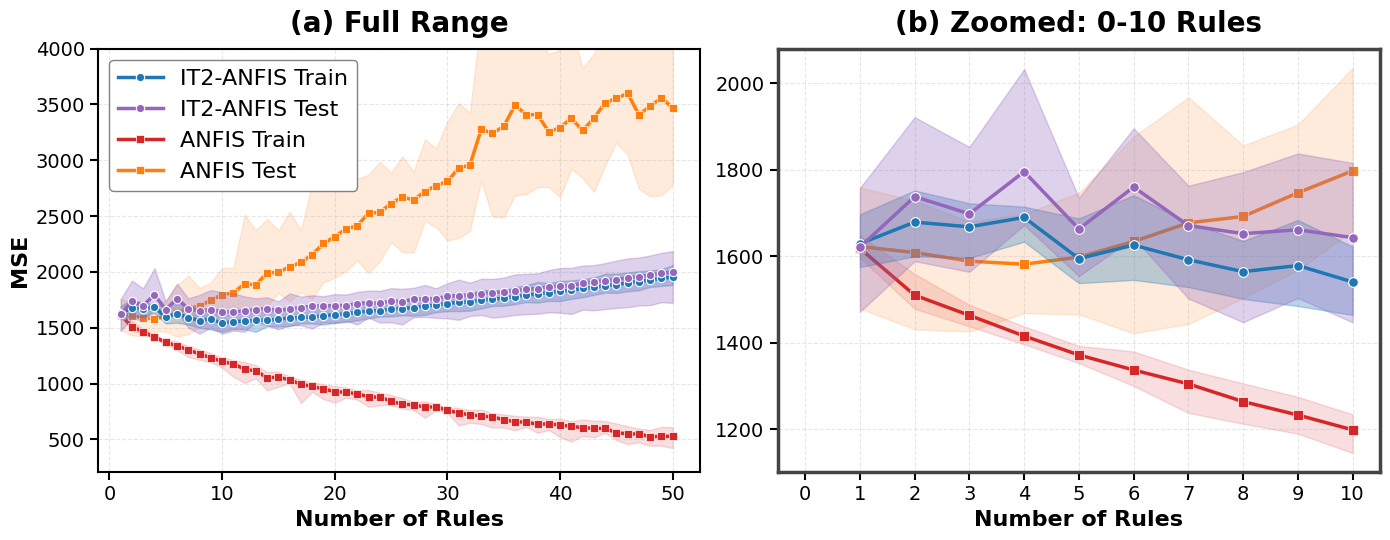}
    \caption{Test performance of ANFIS and IT2-ANFIS across 5–50 rules. (a) Full range; (b) zoomed view of the 0–10 rule region. Each configuration is trained 10 times; solid lines show mean MSE, shaded regions show min–max. IT2-ANFIS attains reduced variance, with optimal performance at 7–10 rules.}
    \label{fig:results2}
\end{figure}

Figure~\ref{fig:results2}(a) shows the results of parameter sweep from 5 to 50 fuzzy rules for IT2-ANFIS and first-order ANFIS, each configuration trained 10 times with different random seeds; panel (b) zooms the critical 5–10 rule region. IT2-ANFIS exhibits a clear U-shaped test MSE curve with a stable optimum in the 7–10 rule range (test MSE $\approx$ 1600 MWh). Beyond 10 rules, test error increases (1970 MWh at 50 rules), indicating overfitting despite regularization. Narrow shaded regions reflect low run-to-run variance and reproducible training. By contrast, first-order ANFIS shows larger variance across seeds and earlier overfitting; the zoomed view highlights poor reproducibility in the low-rule regime. Table~\ref{tab:model_performance} reports additional baselines: Random Forest attains the lowest test error (MSE 1242.28, RMSE 35.25 MWh), while SVM performs poorly (RMSE 51.45 MWh), likely due to the complex, multi-modal relationships in WWTP energy data. Despite RF’s superior point accuracy, its ensemble structure lacks rule-level transparency and interval uncertainty, which are central to this study’s objective. We select the 7-rule IT2-ANFIS for subsequent uncertainty analyses because it (i) achieves near-optimal predictive performance, (ii) provides compact coverage of the input space, and (iii) preserves interpretability and stable training necessary for explainable uncertainty quantification. Further experiments indicate that MSE could be reduced to $\sim$1400 MWh for IT2-AFNIS with careful hyperparameter tuning; however, a systematic optimization is beyond this study’s scope, as the primary focus of the subsequent analysis is uncertainty quantification rather than absolute predictive optimality.

\begin{table}[!htbp]
\centering
\scriptsize
\caption{Performance comparison of different models on the dataset.}
\label{tab:model_performance}
\begin{tabular}{l@{\hskip 0.3cm}c@{\hskip 0.6cm} c S[table-format=2.2] S[table-format=2.2] S[table-format=2.2]}
\toprule
\textbf{Model} & & \textbf{MSE} & \textbf{RMSE} & \textbf{MAE} & \textbf{MAPE (\%)} \\
\midrule
ANFIS-0          & & 1776.95 & 42.15 & 32.26 & 12.61 \\
ANFIS-1          & & 1677.46 & 41.10 & 32.08 & 12.34 \\
IT2-ANFIS   & & 1671.21 & 41.10 & 32.07 & 12.34 \\
RF             & & 1242.28 & 35.25 & 26.63 & 10.38 \\
SVM            & & 2646.85 & 51.45 & 38.70 & 14.83 \\
\bottomrule
\end{tabular}
\end{table}

\subsection{Explainability and Uncertainty Quantification} 
\begin{figure}[!htbp]
    \centering
    \includegraphics[width=0.9\columnwidth]{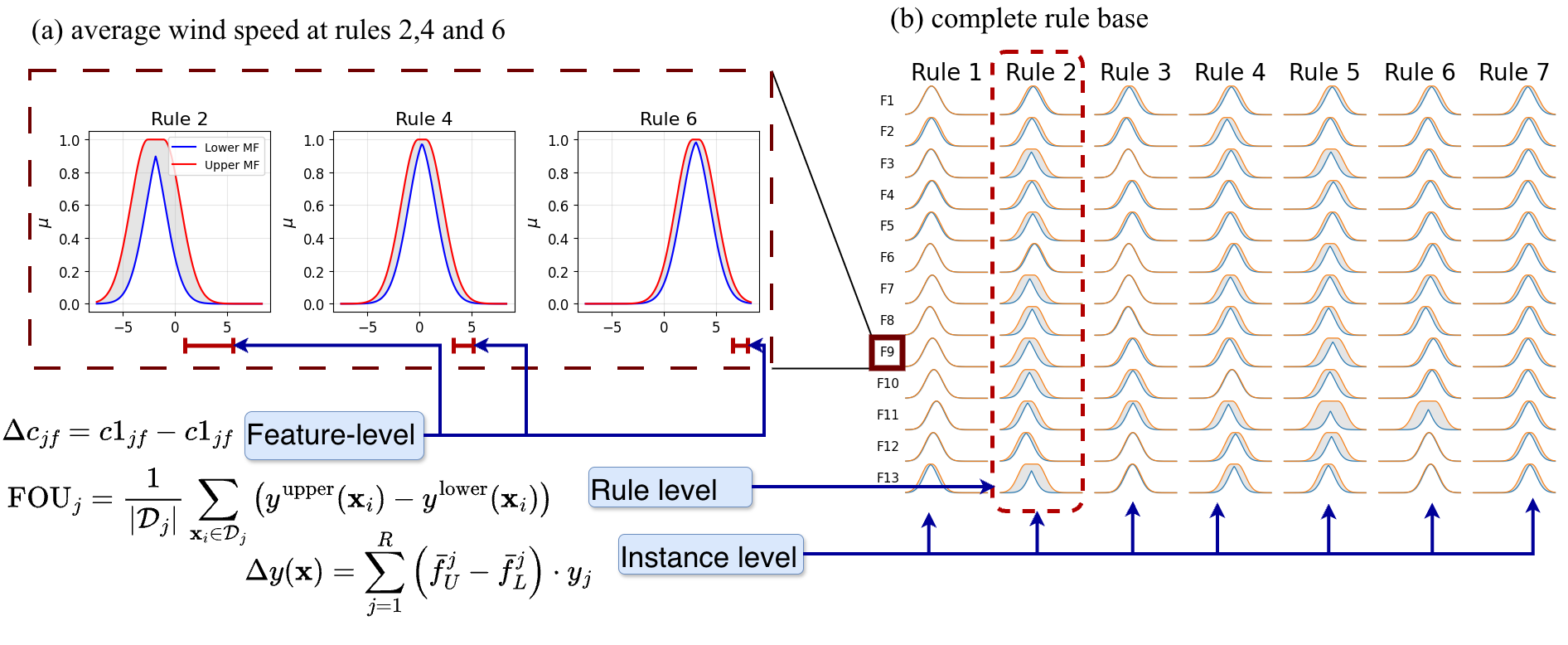}
    \caption{Interval Type-2 Gaussian membership functions learned by the model, with uncertainties depicted at three levels, feature, rule and instance level. (a) highlights three learned rules for average wind speed. (b) presents the complete rule base for all features F1-13.}
    \label{fig:explain}
\end{figure}

While RF can outperform IT-2 in predictive accuracy, it operates as black-box ensembles and lack both uncertainty quantification and rule-level interpretability, which are critical for deployment in safety-critical infrastructure. The proposed IT2-ANFIS framework provides explainable uncertainty quantification at three levels: feature, rule, and instance. Figure \ref{fig:explain} illustrates the type-2 learned interval membership functions in the context of the fuzzy rule base, showing uncertainty represented by FOU bounds. Figure 4.a shows three selected rules with distinct FOU widths. At feature level, average wind speed is represented, showing that rule 4 exhibits a narrow FOU, indicating high confidence and strong data support under moderate operating conditions. In contrast, Rule 2 shows a substantially wider FOU, reflecting higher epistemic uncertainty due to limited or inconsistent data coverage in the medium–low wind speed range. At the rule level, comparing different rules gives an indication of which rule has large uncertainty and which rule holds high confidence. For instance, when comparing different rules in Fig 4.b, rule 2 holds larger uncertainty overall than rule 1 across all 13 inputs features (F1-13). At the instance level, uncertainty is propagated through the firing strengths of activated rules, yielding an interval-valued prediction that represents the cumulative uncertainty across features and rules.

\subsection{Implications of Explainable UQ}
By providing explainable UQ, the framework directly enables practical operational use cases. Bearing explainable uncertainty quantification capabilities directly enable practical operational use cases. In particular, feature- and rule-level FOUs can support sensor prioritization and targeted data collection: input regions associated with wide FOUs (e.g., Rule 2) indicate where additional measurements or higher sampling frequency would most effectively reduce predictive uncertainty. Similarly, rule-level explanations facilitate alarm and maintenance management by localizing uncertainty to specific operating conditions (e.g., medium–low wind or ambient regimes). Operators can exploit this information to adapt alarm handling policies to prediction reliability: wide prediction intervals can trigger verification or inspection workflows to mitigate false alarms, whereas narrow intervals can justify automated responses. In this manner, interpretability can directly inform monitoring and maintenance decisions.

\section{Conclusion}
By leveraging explainable uncertainty quantification, this study enables tracing of predictive uncertainty to specific model parameters and input features, thereby providing diagnostic insight into the modeling pipeline. Using IT2-ANFIS, we predict energy consumption in a large-scale wastewater treatment plant while simultaneously expressing uncertainty in a structured and interpretable manner. This allows identification of operating regimes in which predictions are reliable and those in which they remain uncertain, as well as attribution of these uncertainties to their underlying sources. Such capability highlights the operational conditions and input variables that require richer or higher-quality data to improve model confidence. While the proposed framework primarily addresses epistemic uncertainty, future work will focus on extending the approach to quantify aleatoric uncertainty by integrating fuzzy–probabilistic techniques.

\begin{credits}
\subsubsection{\discintname}
The authors have no competing interests to declare that are
relevant to the content of this article. 
\end{credits}
%
%
%
%

\bibliographystyle{splncs04}
\bibliography{ref}

\end{document}